\pdfoutput=1

\documentclass[11pt]{article}

\usepackage[]{acl}

\usepackage{times}
\usepackage{latexsym}
\usepackage{booktabs}
\usepackage{amsmath}
\usepackage{multirow}
\usepackage{hyperref}
\usepackage[normalem]{ulem}
\useunder{\uline}{\ul}{}
\usepackage[pdftex]{graphicx}
\usepackage{tabularx}
\usepackage{xspace}
\usepackage{algorithmicx}
\usepackage{algorithm}
\usepackage{algpseudocode}
\usepackage{dsfont}
\usepackage{amssymb}
\usepackage{graphicx}

\newcolumntype{L}{>{\scriptsize}l}

\usepackage[T1]{fontenc}

\usepackage[utf8]{inputenc}

\usepackage{microtype}

\usepackage{inconsolata}

\title{Beyond Ranked Lists: The SARAL Framework for Cross-Lingual Document Set Retrieval}

\author{Shantanu Agarwal, Joel Barry, Elizabeth Boschee, Scott Miller\\
        Information Sciences Institute \\ University of Southern California \\ \texttt{\{shantanu,joelb,boschee,smiller\}@isi.edu}}

\begin{document}
\maketitle

\begin{abstract}
Machine Translation for English Retrieval of Information in Any Language (MATERIAL) is an IARPA initiative\footnote{https://www.iarpa.gov/research-programs/material} targeted to advance the state of cross-lingual information retrieval (CLIR).
This report provides a detailed description of Information Sciences Institute's (ISI's) \textbf{S}ummarization and domain-\textbf{A}daptive \textbf{R}etrieval
\textbf{A}cross \textbf{L}anguage's (SARAL's) effort for MATERIAL.
Specifically, we outline our team's novel approach to handle CLIR with emphasis in developing an approach amenable to retrieve a query-relevant document \textit{set}, and not just a ranked document-list.
In MATERIAL’s Phase-3 evaluations, SARAL exceeded the performance of other teams in five out of six evaluation conditions spanning three different languages (Farsi, Kazakh, and Georgian).

\end{abstract}
\section{Introduction}

SARAL's CLIR system consists of four building blocks: a) multiple evidence generators, b) evidence combiner, c) query-document relevance machine, and d) thresholder. 
An evidence generator computes the probability of an English word to be relevant to a foreign sentence. 
Inputs from multiple evidence generators are aggregated using the evidence combiner. 
Using the sentence level probabilities from the evidence combiner, the query-document relevance machine predicts the probability that a foreign document is relevant to an English query. 
The last stage of the pipeline, thresholder, takes the query-document relevance probabilities and decides which set of documents are relevant. 
\section{System components}
In this section, we describe each component of our system in detail.
\subsection{Evidence generator}
\label{subsect:eg}

A canonical approach to CLIR is to map retrieval into a monolingual task by either translating the foreign text, or the query, or both. Limitations of such an approach are that a) the performance is limited to how good the machine translation system is and b) a single machine translation decode is usually insufficient to capture synonymy/polysemy.

To circumvent these issues, we frame the task in a probabilistic framework and start by estimating $p(rel|s_f, w_e)$, the probability that an English word, $w_e$, is related to a foreign sentence, $s_f$. We get this estimate from multiple sources of evidence such as word aligners and various machine translation systems. In addition to these traditional tools, we also leverage SEARCHER \cite{barry-etal-2020-searcher}; a neural framework proposed and developed by our team during the MATERIAL program. We now describe SEARCHER and how we calculate $p(rel| s_f, w_e)$ for each of the evidence sources:

\textit{- SEARCHER}: Different from translating the foreign language sentence or the query, Shared Embedding Architecture for Effective Retrieval (SEARCHER) projects both the English and the foreign text into a shared embedding space. This allows the IR to leverage the dense vector space topology and offers potential advantages in handling synonymy, i.e. where synonymous English terms can match a single foreign term (or vice-versa), as well as for foreign-language polysemy, i.e. where a particular sentence term can have one of several meanings depending on context. 

To be able to obtain $p_{ser}(rel| s_f, w_e)$ from SEARCHER, we leverage parallel corpus such as those used to train machine translation systems. For a bitext sentence pair, $(s_f, s_e)$, any word that occurs in the English sentence, $w_e \in s_e$, is assumed to be relevant to $s_f$ and all other words in a pre-determined fixed English vocabulary, $V_e$, are assumed to be not relevant. 

\begin{figure}[ht!]
\includegraphics[width=8.0cm]{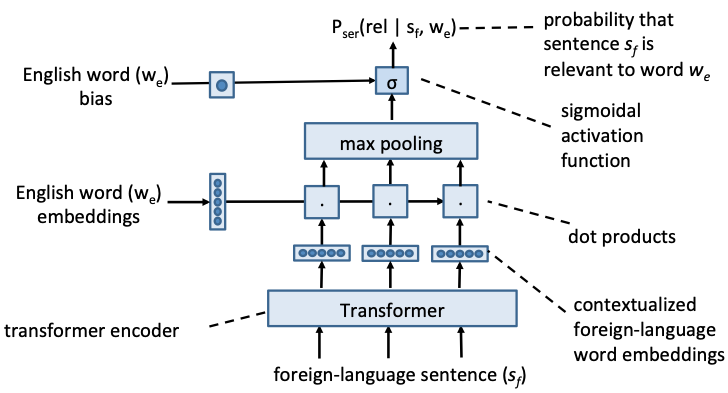}
\caption{Architecture diagram of SEARCHER.}
\label{fig:tsearcher}
\end{figure}  

The architecture diagram for SEARCHER is shown in Fig. \ref{fig:tsearcher}. Contextualized foreign language word embeddings are obtained by feeding $s_f$ through a transformer stack\footnote{We initialize the transformer parameters with pre-trained multilingual embeddings such as XLM-RoBERTa \cite{conneau-etal-2020-unsupervised}.}. 
We compute the dot product between the English word embedding for $w_e$ and each contextualized embedding, taking the maximum dot-product value over all contextualized embeddings.
That value along with a bias term is then passed through a sigmoid function from which we finally get $p_{ser}(rel| s_f, w_e)$. To train SEARCHER, we use the usual cross-entropy loss:
\begin{equation}
\label{eq:cross-entropy}
\begin{aligned}
l = & \sum_{w_e \in s_e} \log p_{ser}(rel| s_f, w_e) \\
&+ \sum_{w_e \in V_e, w_e \notin s_e } \log ( 1- p_{ser}(rel| s_f, w_e) )
\end{aligned}
\end{equation}

\textit{- Word aligners}: We use mGIZA and/or Berkeley word aligners \cite{gao-vogel-2008-parallel, petrov-klein-2007-improved} to get forward word translation probabilities, $p(w_e|w_f)$, using which we get the relevance probability as \footnote{`tt' in $p_{tt}(rel|s_f, w_e)$ stands for translation tables.}:
\begin{equation}
\label{eq:ttref}
p_{tt}(rel|s_f, w_e) = \max_{w_f \in s_f} p(w_e|w_f).
\end{equation}

For speech, an additional complication arises because transcriptions are never perfect. Uncertainties in the word utterances are probabilistically modeled using confusion networks. For speech, we modify Eq. \ref{eq:ttref} as:
\begin{equation}
\label{eq:cnet}
p_{tt}(rel|s_f, w_e) = \max_{w_f \in s_f} p(w_e|w_f) p(w_f),
\end{equation}
where $p(w_f)$ comes from the confusion network. 

\textit{- Machine translation}: 
We run several diverse MT systems in parallel (including a convolutional neural net-based system, and several transformer-based systems), combining their outputs to produce probability estimates for individual terms.

\begin{figure}[ht!]
\centering
\includegraphics[width=7.0cm]{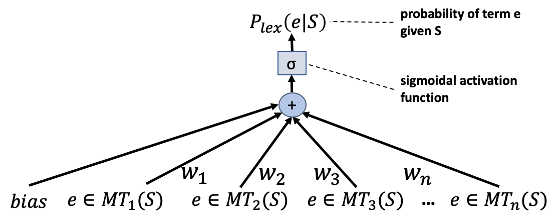}
\caption{Combining multiple MT systems.}
\label{fig:ensembler}
\end{figure}

Specifically, each MT system is associated with a binary feature.
To determine whether a particular term is present, each system sets its feature to “true” if the term occurs in its translation and “false” otherwise. 
These feature values are then combined via logistic regression to determine the probability of the term’s presence, as shown in Figure \ref{fig:ensembler}.

\subsection{Evidence combiner}
\label{subsect:ec}

\begin{figure}[h!]
\centering
\includegraphics[width=7.0cm]{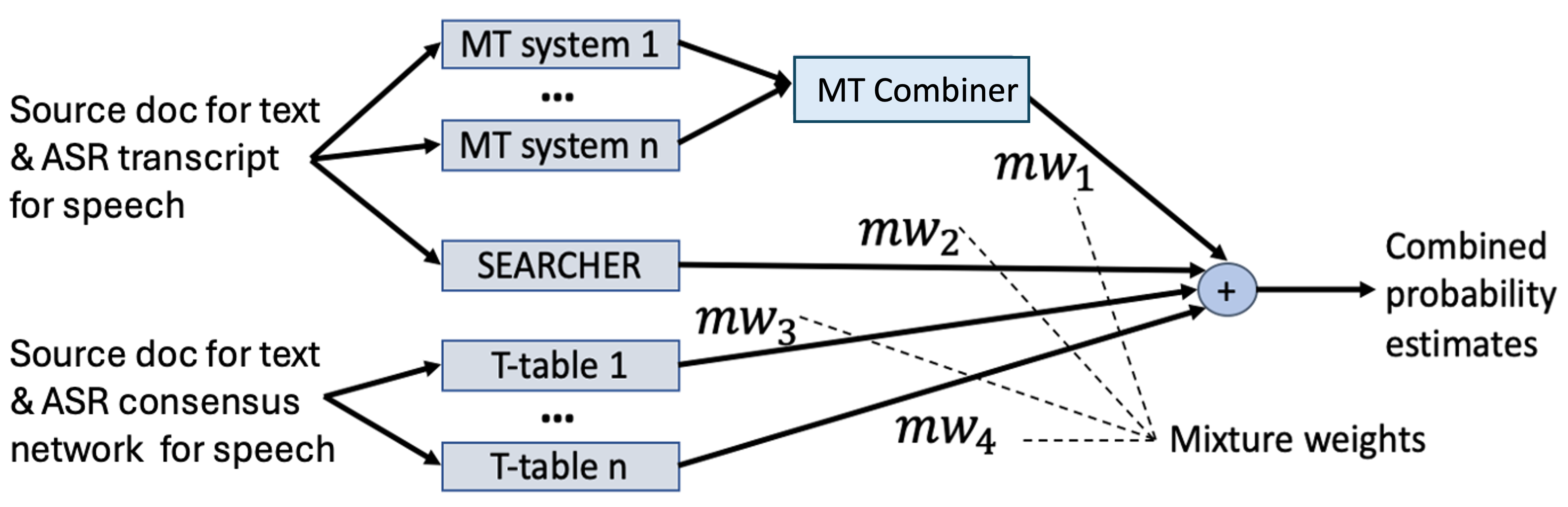}
\caption{Combining evidences from SEARCHER, multiple MT systems, and multiple translation-tables.
}
\label{fig:combiner}
\end{figure}

We have described three mechanisms for obtaining lexical probabilities, each with specific strengths and limitations. 
SEARCHER provides a high degree of diversity; it assigns a probability to every target term given a source sentence. 
But it is insensitive to target language context (although source context is considered). 
The multiple MT systems provide a high degree of accuracy and consider both source and target contexts. 
But diversity is limited by the number of translations available. 
Like SEARCHER, translation tables provide substantial diversity, and additionally facilitate the use of ASR consensus networks for speech, as in Eq. \ref{eq:cnet}. However, neither source nor target context is considered. 
Ultimately, we wish to combine the respective strengths of all these mechanisms to produce the best single estimate of which words are present. 
To do so, we construct a mixture model and estimate the mixture-weights using the EM algorithm as shown in Figure \ref{fig:combiner}.

\subsection{Query-document relevance machine}
\label{subsect:qdrm}
An IR system needs a way to rank the documents based on their relevance to a query. To be able to rank, systems assign a score of relevance to every query-document pair. For the purpose of ranking, it is sufficient for the scores to be just real numbers, but in our system, we constrain the scores to be actual probabilities, $p(rel| d_f, q_e)$, where $d_f$ is the document in foreign language and $q_e$ is the English query. We now show how we leverage $p(rel | s_f, w_e)$ from the evidence combiner to calculate query-document relevance probabilities, $p(rel| d_f, q_e)$. 

MATERIAL supports multiple query types (see Appendix \ref{appendix:query_type}). 
In this report, we focus on SARAL’s approach for handling the most common type: \textbf{lexical queries}. 
A lexical query may consist of multiple phrases and each phrase can consist of multiple words. For a document to be relevant to a lexical query, it must contain a translation equivalent to every word or phrase in the query. For example, $q_e = $ ``scientific research", ``vaccination". For a document to be relevant to such a query, the document needs to be relevant to each query-phrase $(qp_e)$, i.e.
\begin{equation}
\label{eq:pdq}
p(rel | d_f, q_e) = \prod\limits_{qp_e \in q_e} p(rel | d_f, qp_e).
\end{equation}
Now, the probability that a document is relevant to a query-phrase is the union of the probability that the query-phrase is relevant to any of the sentences in the document:
\begin{equation}
\begin{aligned}
\label{eq:pdqp}
p(rel | d_f, qp_e) &= \bigcup_{s_f \in d_f} p(rel | s_f, qp_e) \\
                   &= 1 - \prod\limits_{s_f \in d_f} (1 - p(rel | s_f, qp_e)).
\end{aligned}
\end{equation}
For a query-phrase to be relevant to $s_f$, each word in the phrase, $w_e \in qp_e$, needs to be relevant to $s_f$:
\begin{equation}
\label{eq:psqp}
p(rel | s_f, qp_e) = \prod\limits_{w_e \in qp_e} p(rel | s_f, w_e).
\end{equation}
The right hand side of Eq. \ref{eq:psqp} can be calculated by plugging the values of $p(rel | s_f, w_e)$ from the evidence combiner step. Subsequently, we see that the query-document relevance probability, $p(rel | d_f, q_e)$, can be easily calculated using Eqs. \ref{eq:pdq}-\ref{eq:psqp}. 

\subsection{Thresholder}
\label{subsect:thresholder}
Typical IR systems are judged by how good they are at ranking documents and are evaluated using metrics such as Mean Average Precision (MAP), normalized Discounted Cumulative Gain (nDCG), etc. Different from this traditional approach, the evaluation metric in MATERIAL depends on precision and recall of the set of documents which the system must return for each query. Specifically, systems are evaluated using $mAQWV$\footnote{  \href{https://www.nist.gov/system/files/documents/2017/10/26/aqwv_derivation.pdf}{Original derivation} from NIST.} (modified Actual Query Weighted Value) defined as:
\begin{equation}
\label{eq:aqwv}
mAQWV = \frac{1}{N_q}\sum_{q} QV(q_e).
\end{equation}
In Eq. \ref{eq:aqwv}, $N_q$ is the number of queries used for system evaluation and $QV(q_e)$ is the per-query system score, defined as: 
\begin{equation}
\label{eq:qv}
\begin{aligned}
QV(q_e)  &= 1 - (p_{miss} + \beta p_{fa}),  \\
p_{miss} &= (N_r - N_t)/N_r, \\
p_{fa}   &= N_f/(N - N_r),
\end{aligned}
\end{equation}
where $N_r$ is the gold number of documents relevant to the query, $N_t$ is the correct (true) number of documents returned by the system,  $N_f$ is the incorrect (false) number of documents returned by the system and, $N$ is the total number of documents in the evaluation corpus. The scaling factor $\beta$ determines the relative cost of misses verses false alarms.

From the query-document relevance machine, we get a ranked list of documents, $d_f(i)$, ranked by their query-document relevance probabilities, $p_i(rel | d_f(i), q_e)$. For convenience, we denote this probability as $p_i$ and assume that the documents are sorted in descending order of relevance, i.e. $p_{i} \geq p_{i+1}$. We now show that using $p_i$, one can directly optimize $QV(q_e)$ by precisely determining which set of documents to return for each query. 

For a corpus consisting of $N$ documents, one needs to decide which top $k \in [0, N]$ documents to return. For each value of $k$, the expected number of misses is given by:
\[
E_{miss}(k) =
\begin{cases}
\displaystyle \sum_{i=k+1}^{N} p_i, & \text{if } k < N, \\[6pt]
0, & \text{if } k = N
\end{cases}
\]
and the expected number of false alarms is given by:
\[
E_{fa}(k) =
\begin{cases}
\displaystyle \sum_{i=1}^{k} (1 - p_i), & \text{if } k > 0, \\[6pt]
0, & \text{if } k = 0
\end{cases}
\]
These expectations can be recursively computed for every value of $k$, with a single forward pass to compute the expected number of misses and a single backward pass to compute the expected number of false alarms. Similarly, the expected number of relevant documents in the entire collection is given by:
\[
E_{rel} = \sum_{i=1}^{N} p_i
\]
From these values, we estimate the probability of a miss for each choice of $k$:
\[
p_{miss}(k) = \frac{E_{miss}(k)}{E_{rel}}
\]
and the probability of a false alarm:
\[
p_{fa}(k) = \frac{E_{fa}(k)}{N - E_{rel}}
\]
The expected query value (QV) for each $k$ is then:
\[
E_{\text{QV}}(k) = 1 - \big( p_{miss}(k) - \beta \, p_{fa}(k) \big)
\]
Finally, we select the $k$ that maximizes this value:
\[
\arg\max_k E_{\text{QV}}(k)
\]

While the above provides a precise estimate of the expected QV at each cutoff point, we found that introducing a scaling factor to the expected number of relevant documents, as in \cite{miller07_interspeech, karakos-etal-2020-reformulating}, empirically improves performance. Scaling factors of 1.3 or 1.4 yielded the best gains.
\section{Results}\label{sec:results}

The system described in this report was used for all OP2 MATERIAL languages. In Table \ref{tab:results}, we compare our CLIR system's performance (SARAL) versus that of other participating teams. We see that across multiple languages and across text and speech, SARAL outperforms other teams in 5 out of 6 evaluation configurations. This clearly demonstrates the effectiveness of our approach to CLIR. 

\begin{table}[ht!]
    \footnotesize
    \resizebox{\columnwidth}{!}{
\begin{tabular}{cllll}
& Lang.          & Team-A  & Team-B    & SARAL     \\ \hline\hline
\multirow{3}{*}{ Text } 
& 3S (Farsi)     & 0.827  & 0.798 & \textbf{0.842}  \\
& 3C (Kazakh)    & 0.799 & \textbf{0.825}  & 0.785           \\
& 3B (Georgian)  & 0.821 & 0.806 & \textbf{0.850}  \\ \hline
\multirow{3}{*}{ Speech } 
& 3S (Farsi)     & 0.687 & 0.694 & \textbf{0.719}  \\
& 3C (Kazakh)    & 0.705 & 0.697 & \textbf{0.742}      \\
& 3B (Georgian)  & 0.653 & 0.730 & \textbf{0.795}  \\
\end{tabular} 
}
\caption{\small Text and speech $mAQWV$ scores comparing various teams across all OP2 MATERIAL languages. All evaluations were run using a predetermined value of $\beta = 40$.}
\label{tab:results} 
\end{table}

\section*{Acknowledgments}
This research is based upon work supported in part by the Office of the Director of National Intelligence (ODNI), Intelligence Advanced Research Projects Activity (IARPA), via contract\# FA8650-17-C9116.
The views and conclusions contained herein are those of the authors and should not be interpreted as necessarily representing the official policies, either expressed or implied, of ODNI, IARPA, or the U.S. Government. 
The U.S. Government is authorized to reproduce and distribute reprints for governmental purposes notwithstanding any copyright annotation therein.

\bibliography{anthology,custom}

\begin{thebibliography}{6}
\expandafter\ifx\csname natexlab\endcsname\relax\def\natexlab#1{#1}\fi

\bibitem[{Barry et~al.(2020)Barry, Boschee, Freedman, and Miller}]{barry-etal-2020-searcher}
Joel Barry, Elizabeth Boschee, Marjorie Freedman, and Scott Miller. 2020.
\newblock \href {https://aclanthology.org/2020.clssts-1.4} {{SEARCHER}: Shared embedding architecture for effective retrieval}.
\newblock In \emph{Proceedings of the workshop on Cross-Language Search and Summarization of Text and Speech (CLSSTS2020)}, pages 22--25, Marseille, France. European Language Resources Association.

\bibitem[{Conneau et~al.(2020)Conneau, Khandelwal, Goyal, Chaudhary, Wenzek, Guzm{\'a}n, Grave, Ott, Zettlemoyer, and Stoyanov}]{conneau-etal-2020-unsupervised}
Alexis Conneau, Kartikay Khandelwal, Naman Goyal, Vishrav Chaudhary, Guillaume Wenzek, Francisco Guzm{\'a}n, Edouard Grave, Myle Ott, Luke Zettlemoyer, and Veselin Stoyanov. 2020.
\newblock \href {https://doi.org/10.18653/v1/2020.acl-main.747} {Unsupervised cross-lingual representation learning at scale}.
\newblock In \emph{Proceedings of the 58th Annual Meeting of the Association for Computational Linguistics}, pages 8440--8451, Online. Association for Computational Linguistics.

\bibitem[{Gao and Vogel(2008)}]{gao-vogel-2008-parallel}
Qin Gao and Stephan Vogel. 2008.
\newblock \href {https://aclanthology.org/W08-0509} {Parallel implementations of word alignment tool}.
\newblock In \emph{Software Engineering, Testing, and Quality Assurance for Natural Language Processing}, pages 49--57, Columbus, Ohio. Association for Computational Linguistics.

\bibitem[{Karakos et~al.(2020)Karakos, Zbib, Hartmann, Schwartz, and Makhoul}]{karakos-etal-2020-reformulating}
Damianos Karakos, Rabih Zbib, William Hartmann, Richard Schwartz, and John Makhoul. 2020.
\newblock \href {https://aclanthology.org/2020.clssts-1.7} {Reformulating information retrieval from speech and text as a detection problem}.
\newblock In \emph{Proceedings of the workshop on Cross-Language Search and Summarization of Text and Speech (CLSSTS2020)}, pages 38--43, Marseille, France. European Language Resources Association.

\bibitem[{Miller et~al.(2007)Miller, Kleber, Kao, Kimball, Colthurst, Lowe, Schwartz, and Gish}]{miller07_interspeech}
David R.~H. Miller, Michael Kleber, Chia-Lin Kao, Owen Kimball, Thomas Colthurst, Stephen~A. Lowe, Richard~M. Schwartz, and Herbert Gish. 2007.
\newblock \href {https://doi.org/10.21437/Interspeech.2007-174} {Rapid and accurate spoken term detection}.
\newblock In \emph{Interspeech 2007}, pages 314--317.

\bibitem[{Petrov and Klein(2007)}]{petrov-klein-2007-improved}
Slav Petrov and Dan Klein. 2007.
\newblock \href {https://aclanthology.org/N07-1051} {Improved inference for unlexicalized parsing}.
\newblock In \emph{Human Language Technologies 2007: The Conference of the North {A}merican Chapter of the Association for Computational Linguistics; Proceedings of the Main Conference}, pages 404--411, Rochester, New York. Association for Computational Linguistics.

\end{thebibliography}

\appendix

\section{Query types in MATERIAL}
\label{appendix:query_type}

In the MATERIAL program, the performers need to develop system to serve 3 distinct types of queries\footnote{  \href{https://www.nist.gov/system/files/documents/2018/07/12/openclirqueriesandrelevance.pdf}{Query definitions} from MATERIAL.}:
\begin{itemize}
\item{Lexical queries: The goal of lexical queries is to find documents containing a translation equivalent of the query expression.
These single queries could be a single phrase, such as $``\text{campaign}"$, and $``\text{pet cat}"$, or have multiple phrases, such as $phrase_1 = ``\text{brother}"$, $phrase_2 =``\text{auction}"$.
}
\item{Conceptual queries: The goal of these queries is to find documents with content relevant to the topic introduced by the query term.
For example, $``\text{safari+}"$, in which case the system should retrieve documents pertaining to $``\text{big game hunting expeditions in
East Africa}"$ but not to $``\text{a cruise along the Nile}"$.
}
\item{EXAMPLE\_OF queries: These queries aim to retrieve documents that mention a specific example of the query term.
For instance, for the query $``\text{EXAMPLE\_OF(virus)}"$, the system should return documents referring to viruses such as $``\text{HIV}"$ or $``\text{influenza}"$, while avoiding documents that mention non-viral diseases.
}
\end{itemize}

In the MATERIAL evaluations, the query set included a mixture of all the above query types. 
However, ``Lexical queries" were the most common, and this report focuses primarily on SARAL’s approach for handling them.

\end{document}